\documentclass[10pt,twocolumn,letterpaper]{article}

\usepackage[final]{cvpr}      

\usepackage{graphicx}
\usepackage{amsmath}
\usepackage{amssymb}
\usepackage{booktabs}
\usepackage{multirow}
\usepackage{lipsum}
\usepackage{tabularx}
\usepackage{array}
\usepackage{longtable}

\usepackage[pagebackref,breaklinks,colorlinks]{hyperref}

\usepackage[capitalize]{cleveref}
\crefname{section}{Sec.}{Secs.}
\Crefname{section}{Section}{Sections}
\Crefname{table}{Table}{Tables}
\crefname{table}{Tab.}{Tabs.}

\begin{document}

\title{NOVI : Chatbot System for University Novice  with BERT and LLMs}

\author{Yoonji Nam\\
Sungkyunkwan University\\
Department of Data Science Convergence\\
{\tt\small seinundzeit@g.skku.edu}
\and
TaeWoong Seo\\
Sungkyunkwan University\\
Department of Artificial Intelligence Convergence\\
{\tt\small tae0204@g.skku.edu}
\and
Gyeongcheol Shin\\
Sungkyunkwan University\\
Department of Classical Chinese Literature\\
{\tt\small gospel1011@g.skku.edu}
\and
Sangji Lee\\
Sungkyunkwan University\\
Department of Immersive Media Engineering\\
{\tt\small sangji9504@gmail.com}
\and
JaeEun Im\\
Sungkyunkwan University\\
Department of Immersive Media Engineering\\
{\tt\small imje97@skku.edu}
}

\maketitle

\begin{abstract}

To mitigate the difficulties of  university freshmen in adapting to university life, we developed NOVI, a chatbot system based on GPT-4o. This system utilizes post and comment data from SKKU 'Everytime' , a university community site. Developed using LangChain, NOVI's performance has been evaluated with a BLEU score,  Perplexity score, ROUGE-1 score, ROUGE-2 score, ROUGE-L score and METEOR score. This approach is not only limited to help university freshmen but also expected to help various people adapting to new environments with different data. This research explores the development and potential application of new educational technology tools, contributing to easier social adaptation for beginners and settling a foundation for future advancement in LLM studies.

\end{abstract}

\section{Introduction}
\label{sec:intro}

Recently, various enterprises and public institutions, as well as several universities in Korea, have introduced chatbots to provide information related to academic administration and IT services. Examples include Sungkyunkwan University's academic administration chatbot 'Kingobot,' Seoul University's IT service chatbot 'Snubot,' Hanyang University's academic administration chatbot 'Goongmoonyang,' Yonsei University's library materials loan and purchase chatbot 'Tuksoori,' and Chung-Ang University's academic administration chatbot 'CHARLI' \cite{AnSeonghun2022University}. These on-campus chatbots can reduce the workload of staff by addressing frequently asked academic-related inquiries and provide students with necessary information at any time, thereby offering efficient and highly satisfactory services \cite{Hur2020}.

However, university-provided chatbots are rule-based rather than AI-based, which presents a significant limitation. These rule-based chatbots struggle to handle queries that deviate from their predefined rules and scenarios\cite{MinYoungtae2023Library}. For instance, Sungkyunkwan University's 'Kingobot' cannot respond to untrained information queries or only provides information that includes similar keywords (see Figure 1). As of now, chatbots cannot answer specific questions and only offer limited functionalities, posing a significant limitation in providing practical information for students' campus life.

\begin{figure}[h]
    \centering
    \includegraphics[width=0.7\linewidth]{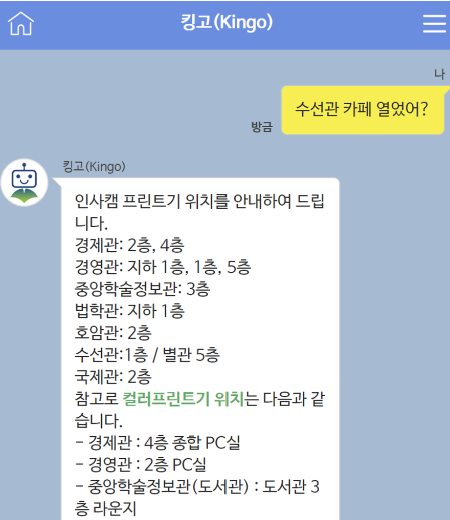}
    \caption{Error of KingoBot}
    \label{fig:enter-label}
\end{figure}

The drawbacks of these chatbots are particularly critical considering the high demand for practical information about universities, especially among freshmen. Freshmen face confusion and difficulties due to changes in learning methods, expanded areas of activity, and new interpersonal relationships upon entering university\cite{JungYoukyung2022Freshmen}. Various studies have shown that freshmen have significantly lower adaptation levels to school life compared to other academic years and experience more negative emotions such as depression and anxiety\cite{JeonBora2017University}. These negative emotions can persistently impact university life, leading to academic underperformance and dropout\cite{kim2009importance}, making it a priority for universities to provide information to help freshmen adapt.

Additionally, the MZ generation, often referred to as the 'Call Phobia' generation, prefers message-based communication over face-to-face or voice communication favored by the X generation. For future generations more accustomed to text-based communication than phone calls, providing information through text-based chatbots will become increasingly important.

Numerous studies on chatbots have explored effective ways to build academic administration chatbots\cite{AnSeonghun2022University} and developed academic guidance chatbot systems utilizing natural language processing\cite{kim2018, Hur2020, ParkHangil2023AI}. However, despite the active research on generative AI chatbots in various fields, the application of the latest methodologies in academic information chatbots remains inadequate. There is a lack of chatbot research that has been developed by learning the practical data needed by freshmen.

In a situation where the academic information provided by the school is limited, university students are increasingly seeking information from online communities such as 'Everytime' instead of the official academic information provided by their universities. Previous research indicates that university students find these online communities useful for seeking and sharing everyday information, with freshmen and sophomores showing higher dependence on 'Everytime' compared to juniors and seniors\cite{LeeByungchan2015Leisure,ChoiSinae2021Online}. Therefore, research on chatbots trained with community database information is necessary.

This study aims to develop a chatbot that provides academic information to freshmen based on an online community database, offering specific information needed for adapting to university life. We expect to provide practical assistance to university freshmen in adjusting to university life and acquiring the necessary information to embark on a successful first year.

\section{Related Work}

\subsection{Chatbot system for university }

\begin{table*}[ht]
\centering
\begin{tabular}{|m{3cm}|m{1cm}|m{11cm}|}
\hline
{Author} &{Year} & {Result} \\ \hline
Park et al. \cite{ParkHangil2023AI} & 2023 & AI-based university information notification chatbot system \\ \hline
An, S. \cite{AnSeonghun2022University} & 2022 & Button, scenario, keyword, AI, hybrid based AI chatbot system structure, SNS based AI chatbot system structure \\ \hline
Min \& Kwak \cite{MinYoungtae2023Library} & 2023 & AI chatbot service in library \\ \hline
Kim et al. \cite{kim2018} & 2018 & NLP, Open Source Software based chatbot dialogue system \\ \hline
Hur et al. \cite{Hur2020} & 2020 & NLP based University Bulletin Chat-bot System \\ \hline
Neupane, S.\cite{neupane2024questions} & 2024 & LLM Based chatbot system for University Resources using RAG pipelines \\ \hline
Tin et al. \cite{lai2023psyllm} & 2023 & Psy-LLM framework (AI-based tool that leverages large language models to handle questions in psychological consultation settings) \\ \hline
\end{tabular}
\caption{Previous research about University Chatbot system}
\end{table*}

In recent years, there has been increasing interest in utilizing AI for enhancing various services in educational institutions, particularly through the development of chatbot systems. Several studies have explored different aspects of implementing AI-driven chatbots to support university operations and services.

Neupane et al. \cite{neupane2024questions} introduced BARKPLUG V.2, a chatbot system based on a Large Language Model (LLM) and built using Retrieval Augmented Generation (RAG) pipelines. This system is aimed at enhancing the user experience and improving access to university information by providing interactive responses about academic departments, programs, campus facilities, and student resources. The study shows impressive quantitative performance with a mean RAGAS score of 0.96 and high user satisfaction as per the System Usability Scale (SUS) surveys.

Kim, Jung, and Kang \cite{kim2018} presented a chatbot system for handling academic inquiries using natural language processing (NLP) and open-source software. This system was designed to alleviate the increased workload on university departments during the influx of inquiries from new students.

Hur et al.\cite{Hur2020} developed a university guide chatbot system based on NLP, utilizing information collected from university regulations and guidebooks to create a dictionary-style database for responses. This system aimed to streamline the process of providing information to students about university policies and campus life.

Park et al. \cite{ParkHangil2023AI} focused on the development of an AI-based university information notification chatbot system. Their study, presented at the KIIT Conference, emphasized the importance of providing timely and accurate information to students through an intelligent chatbot interface.

Choi, Cho, and Kim \cite{ChoiSinae2021Online} designed a chatbot system to improve accessibility to university counseling services during the COVID-19 pandemic. This system aimed to facilitate easier access to support services for students, ensuring they receive timely assistance during challenging times.

An \cite{AnSeonghun2022University} explored strategies for constructing an effective chatbot system to guide university academic operations. The study examined various approaches to designing chatbots that can efficiently handle academic inquiries and provide reliable information to students and staff. This paper proposed a button, scenario, keyword, AI, and hybrid-based chatbot system structure, as well as an SNS-based AI chatbot system structure, to optimize the interaction and functionality of academic chatbots.

These studies collectively highlight the diverse applications of AI-driven chatbots in university settings, demonstrating their potential to improve service delivery, enhance user experience, and alleviate the workload on university staff.

\subsection{Applications in Natural Language Processing (NLP)}

Papineni et al. \cite{papineni2002bleu} introduced the BLEU metric, which evaluates the quality of machine translation by comparing the number of tokens in the generated sentence to the number of tokens in the reference (correct) sentence. This metric has been widely adopted for its straightforward approach in assessing translation accuracy. Lavie and Agarwal \cite{banerjeeLAVIE2005meteor} proposed the METEOR metric to address some of the BLEU metric's limitations. METEOR evaluates translations by calculating a score based on explicit word-to-word matches between the machine translation and the reference translation, considering synonyms and stemming.

In the domain of text summarization, Lin\cite{lin-2004-rouge} developed the ROUGE metric, which is used to evaluate the quality of automatic summaries by focusing on n-gram matching between the generated summary and reference summaries. ROUGE has become a standard evaluation metric for summarization tasks, providing insights into the coherence and coverage of the summaries produced by models.

Recent advancements in spam detection have leveraged transformer-based embeddings and ensemble learning. Ghourabi and Alohaly\cite{ghourabi2023enhancing} conducted a study comparing various spam detection techniques, including traditional methods like TF-IDF and word embeddings against modern transformer-based models such as BERT and GPT. Their findings indicate that transformer-based methods significantly outperform traditional techniques in classifying spam/ham emails. Similarly, AbdulNabi and Yaseen\cite{yaseen2021spam} explored the effectiveness of using BERT for both tokenizer and classifier roles in spam email detection, concluding that BERT-based approaches achieved top performance in this area.

In summary, the evolution of evaluation metrics for machine translation and text summarization, alongside the adoption of advanced language models for spam detection, demonstrates significant progress in NLP research. These developments highlight the continuous improvements in model accuracy and efficiency, paving the way for more robust and reliable AI applications.

\section{Method}

\begin{figure}[h]
    \centering
    \includegraphics[width=1\linewidth]{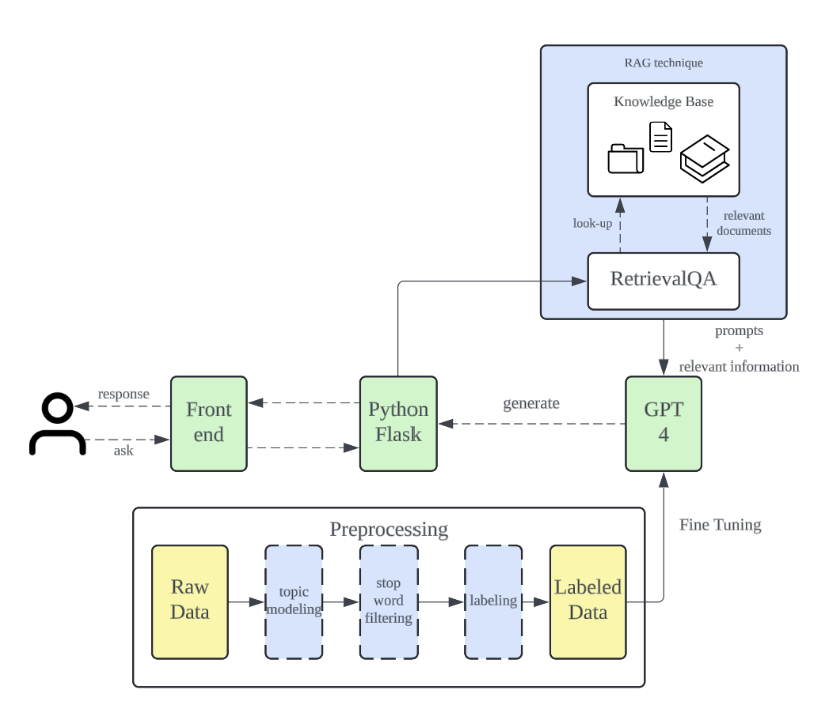}
    \caption{System Overview}

\end{figure}

The overall flow of the system is as above(Figure 2). When a user asks a query, the query is received at the front end and the query is handed over to Flask. Flask delivers the query to RetrievalQA. Then, in the RetrievalQA step, information related to the query is found in the existing trusted knowledge base. The queries and information are then transferred to a pre-fine-tune model, which generates a corresponding answer. The answer is communicated to the user at the front end through Python Flask. In this case, the data for fine-tuning the model is preprocessed as follows.  First, we took Raw Data and did topic modeling to extract only data related to school and freshmen. Then, unnecessary words and swear words are removed through stop word filtering, and labeling is performed based on the significance of information.

Everytime is an online community service designed for the purpose of facilitating the sharing of information within the same university. This platform allows users to select their university and undergo a verification process to confirm their identity. Once verified, users can participate in various discussion boards where they can post questions and receive answers from other members of the university community. Each university on the platform hosts multiple boards, such as information boards, promotion boards, and career and employment boards. Due to the unique characteristics of each board in terms of the types of questions and answers typically found there, it was necessary for us to carefully select the boards that were most appropriate for our project. This selective approach ensured that we accessed the most relevant and useful information for our specific needs.


\subsection{Data Collection}
We collected data from three specific boards on Everytime:

\textbf{Ask Broly}: The Broly board is where users can ask questions related to academics, studies, career paths, and employment to a user nicknamed Broly. This board was the most suitable for our project.

\textbf{Freshmen Board}: The Freshmen Board is where new students can ask questions about school life, and seniors provide information through their answers.

\textbf{Insa Campus Board}: Sungkyunkwan University has two campuses: the Humanities and Social Sciences Campus and the Natural Sciences Campus. Since our project targeted freshmen on the Humanities and Social Sciences Campus, we chose only the Insa Campus Board

\subsection{Data Collection Methods}
To collect data, we utilized web crawling techniques on the Everytime community platform, employing the Selenium library. The crawling process was as follows:
Crawler Setup: Installed ChromeDriver with \textit{webdriver\_manager} and used Selenium's \textit{webdriver.Chrome} for browser automation.

\textbf{Login}: Since Everytime restricts post access to logged-in users, the crawler was programmed to log in automatically by entering the user ID and password into the login form and submitting it.

\textbf{Collecting Post Lists}: The crawler navigated through the pages of each board to collect lists of posts. We set a range of page numbers, accessed each page, and extracted the URLs of the posts.

\textbf{Collecting Detailed Post Information}: The crawler visited each extracted post URL to retrieve detailed information, including the post's title, content, date, number of likes, number of scraps, and comments. Comments were extracted separately and stored in a structured format.

\textbf{Data Storage}: The collected data was transformed into a dataframe using the Pandas library and saved as an Excel file.

\textbf{Error Handling and Logging}: We implemented mechanisms to handle errors during the crawling process, ensuring data was saved up to the point of any error. The crawling process was logged to a file for future review and troubleshooting.

\subsection{Data Filtering}

\begin{figure}[h]
    \centering
    \includegraphics[width=1\linewidth]{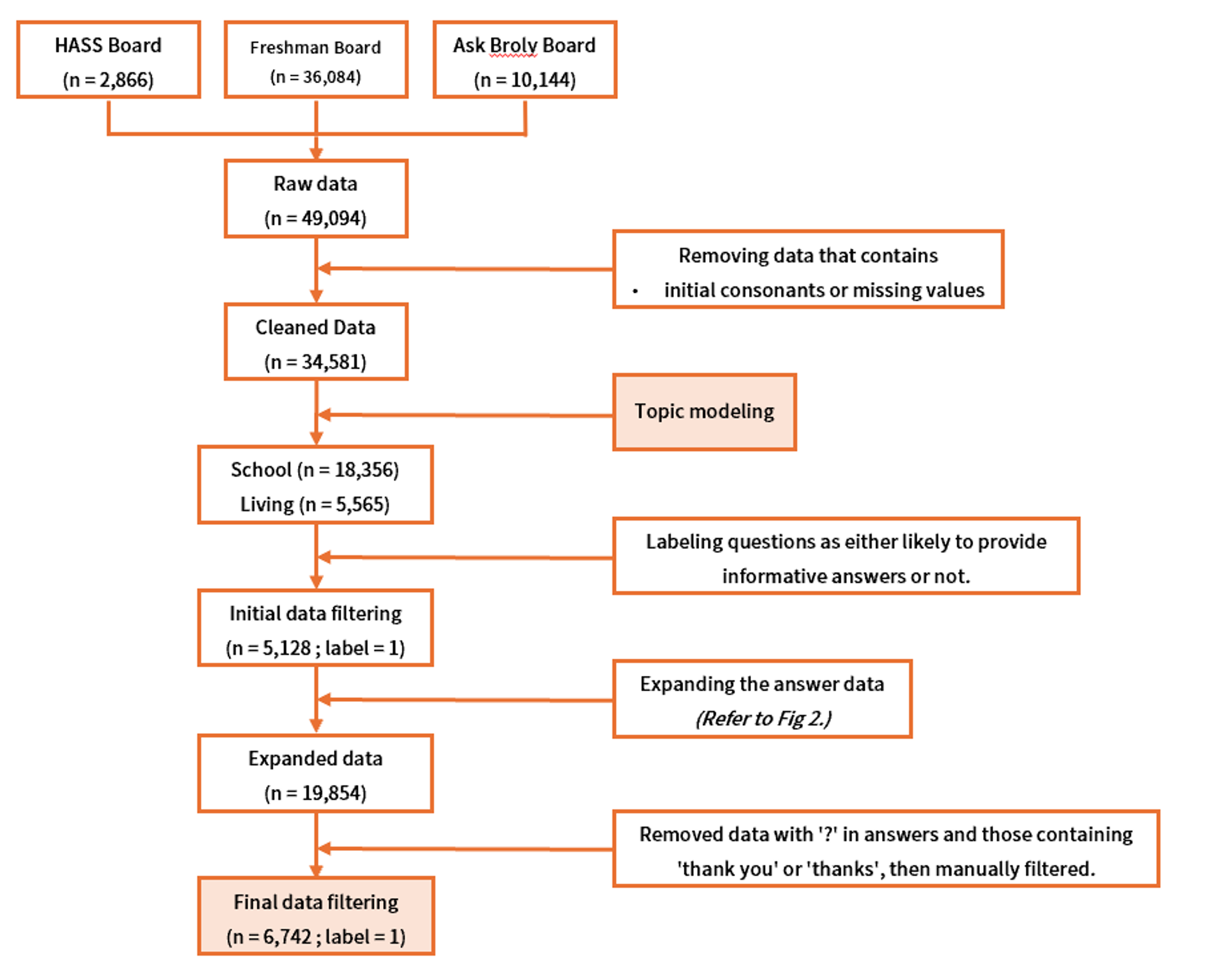}
    \caption{Flowchart of the filtering process}

\end{figure}

The data filtering process, illustrated in figure 3, was conducted as follows:

a. Initial Data Cleaning: During the data integration phase, we removed entries that contained only partial information, such as names with only initials (e.g., "Prof. KJK" rather than "Prof. Jaekwang Kim"), and handled missing values. Entries with only initials were excluded because they make it difficult to accurately identify the specific course being referred to.

b. Topic Modeling: The discussion boards included questions and answers unrelated to school life, such as those about politics, society, and private matters. We applied topic modeling techniques to filter out questions not related to academics. This process reduced the total dataset to 18,356 academic-related entries and 5,565 living-related entries.

c. Manual Labeling: Researchers manually read the questions and labeled them in binary terms to determine whether the answers provided useful information. Due to limitations in human resources, only 12,411 entries were labeled, resulting in 5,128 useful (1) and 7,283 not useful (0) labels.

d. Expanding \& Manual Labeling: As shown in the figure 4, the original dataset contained multiple answers within a list inside a dictionary, which needed to be unfolded. After expanding the 5,128 entries, we removed irrelevant responses such as "Thank you" and "?" and conducted additional manual labeling, resulting in a total of 14,829 useful data entries.

\begin{figure}[h]
    \centering
    \includegraphics[width=1\linewidth]{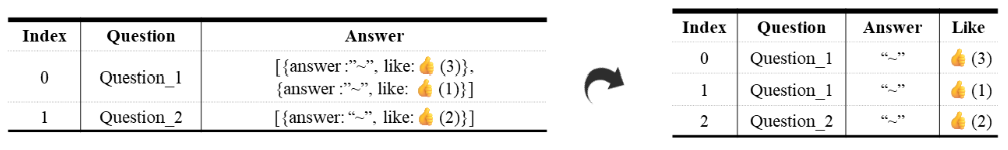}
    \caption{Expand the answers}

\end{figure}

\subsection{Novel Data Filtering Methodology}

After manually labeling the data to determine whether the questions were likely to provide useful information post-topic modeling, we identified inefficiencies in the existing manual labeling process. To address these inefficiencies, we devised a new data filtering methodology inspired by binary classification tasks used in spam/ham email filtering\cite{ghourabi2023enhancing}.

\subsubsection{Tokenization and Embedding }  
We utilized pretrained transformer models, such as GPT and BERT, for tokenizing and embedding the question data. These models were chosen because previous research demonstrated their superior performance.


\textbf{Klue-BERT-base}: KLUE BERT base is a pre-trained BERT model specifically designed for the Korean language. The developers created KLUE BERT base as part of the Korean Language Understanding Evaluation (KLUE) Benchmark initiative to enhance the performance of natural language processing tasks in Korean.

\textbf{GPT}:  GPT-ADA v2 is a state-of-the-art language model developed by OpenAI, designed for efficient and versatile natural language understanding and generation. It excels in various applications such as text completion, summarization, translation, and conversational AI. With improved accuracy and response time, GPT-ADA v2 is a powerful tool for developers and businesses.

\subsubsection{Classification Models}  
Once the data was embedded using BERT and GPT, we employed several classification models, including Support Vector Machines (SVM), LightGBM, and Convolutional Neural Networks (CNN), to perform the classification tasks. We chose the three models because they demonstrated the highest performance in previous studies.

\subsubsection{Model Comparison}  

We compared the performance of the three best-performing models to classify the dataset, which consisted of 5,128 useful (1) and 7,283 not useful (0) labeled entries.

 By implementing this methodology, future manual labeling tasks can be significantly streamlined, as this approach provides a clear framework to follow, enhancing both the efficiency and effectiveness of the filtering process.

\begin{figure}[h]
    \centering
    \includegraphics[width=\linewidth]{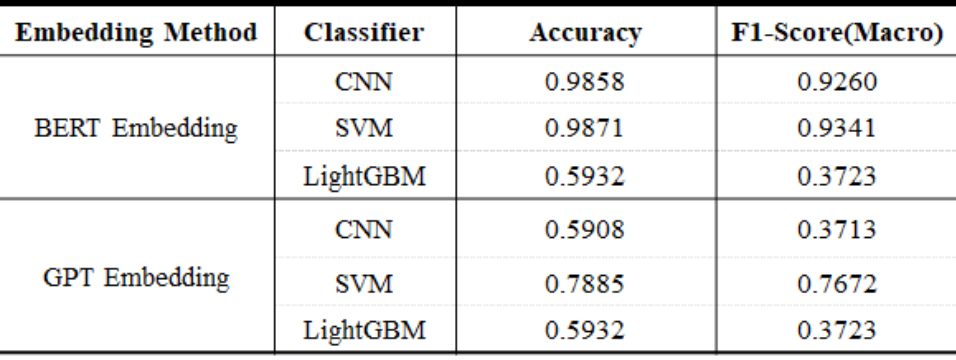}
    \caption{Classification results}

\end{figure}

We expected the model to produce results similar to those filtered by humans. Therefore, we tested 100 completely human-generated test cases using the BERT + CNN model, but the accuracy was only 0.686. Consequently, we decided to train the LLM using data that we labeled ourselves this time. By adjusting hyperparameters and implementing this method in the future, manual labeling tasks can be significantly streamlined, as this approach provides a clear framework to follow, enhancing both the efficiency and effectiveness of the filtering process.

\subsection{LLM}

\subsubsection{Introduction of LLM}
Large Language Models (LLMs) are newly emerging tools in natural language processing (NLP) that perform functions such as understanding human language and generating corresponding responses by training on large-scale datasets. These models are designed based on the Transformer architecture and consist of trillions of parameters. Currently, prominent models include GPT-4 and Llama3. In this study, we used one of the LLMs, the GPT-4o model, for training and implemented the framework using Langchain. This enabled us to develop a system capable of generating accurate and efficient responses to natural language questions.

\subsubsection{Components of Langchain}
Langchain is an open-source framework that facilitates the efficient construction of LLM-based applications. Langchain provides various components to easily build and manage natural language processing systems. The following are descriptions of each tool:

\begin{table}[h]
    \centering
    \begin{tabular}{|c|p{5cm}|c|}
    \hline
    Tool              & Description                                                                                                                                                    \\ \hline
    RetrievalQA       & Provides a function to search and retrieve information stored in the Vector Store based on user queries.                                                       \\ \hline
    QA Chain          & A chain that generates the final response based on the retrieved data.                                                                                         \\ \hline
    Vector Store(RAG) & A structure for storing embedded text data and efficiently searching through it. In this study, we used FAISS.                                                 \\ \hline
    Flask             & A web application that provides a user interface, processes queries, and displays the results to the user.                                                     \\ \hline
    Embeddings        & Tokenizes and embeds text data loaded from CSVLoader, converting it into a format that the model can process. In this study, we used OpenAI's embedding model. \\ \hline
    Prompt Template   & A template that specifies the format for generating responses based on the input data from the user.                                                           \\ \hline
    CSV Loader        & A tool for loading data files, embedding them, storing them in the Vector Store.                                                                               \\ \hline
    \end{tabular}
    \caption{Tool description}
    \label{tab:my_label}
\end{table}

The system built using Langchain operates as follows:
1. The user accesses the URL created with Flask.
2. The user enters a question.
3. The entered question is tokenized and embedded.
4. The embedded question is used to search for relevant data in the Vector Store.
5. The retrieved data is passed to the QA chain to generate the final response.
6. The generated response is returned to the user via Flask.

\begin{figure}[h]
    \centering
    \includegraphics[width=1\linewidth]{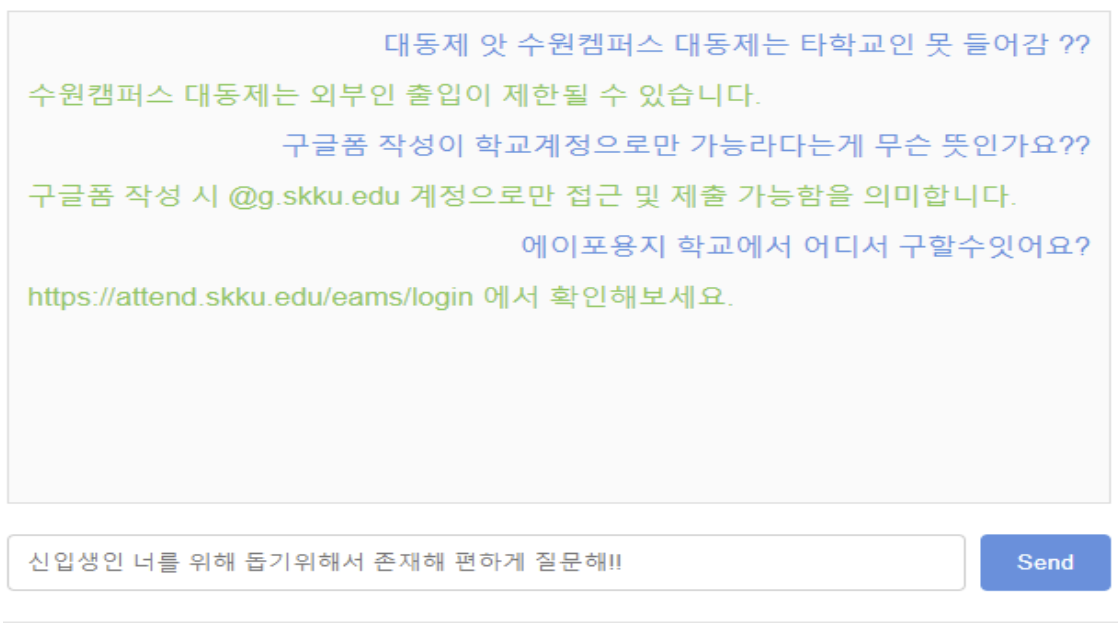}
    \caption{ChatBot Simulation}

\end{figure}
The model understood Korean well and was able to provide high-quality answers specialized in information about Sungkyunkwan University.

\section{Results}

\subsection{Performance Evaluation Metrics and Test Case Creation }

In this study, the performance of the LLM was evaluated using the following metrics: The evaluation metrics used were BLEU, Perplexity, ROUGE, and METEOR.

\subsubsection{BLEU (Bilingual Evaluation Understudy)}

\textbf{Features}: Used to evaluate the quality of machine translation. Measures the lexical similarity between the reference translation and the generated translation. Here, BP is the brevity penalty, $\alpha$ is the n-gram weight, and $P_n$ is the n-gram precision.
BLEU score is an indicator mainly used to evaluate the quality of machine translation. It determines the suitability of sentences generated by generative language models by measuring the degree of agreement of words or phrases between machine translation results and human translation results. BLEU's equation is as follows:

\begin{equation}
\text{BLEU} = BP \cdot \exp \left( \sum_{n=1}^N \alpha_n \log P_n \right)
\end{equation}

\subsubsection{ROUGE (Recall-Oriented Understudy for Gisting Evaluation)}

\textbf{Features}: Primarily used to evaluate the quality of text summarization. Measures the overlap between the reference summary and the generated summary. Here, $\text{Count}_{\text{match}}$ is the number of matching n-grams between the reference summary and the generated summary, and $\text{Count}$ is the number of n-grams in the reference summary.

The ROUGE score is the Recall-Oriented Understudy for Gisting Evaluation, which is an indicator of the performance evaluation of the text summarization model. ROUGE mainly evaluates task performance such as summarization and machine translation, and calculates scores by comparing summarized or translated sentences generated by models with human-made references. As the name suggests, ROUGE calculates scores using Recall and Precision. In general, it can be viewed as a metric evaluated based on n-gram recall. The equation of ROUGE-N for N-gram is as follows:
\begin{equation}
\text{ROUGE-N} = \frac{\sum_{s \in \text{Ref Summaries}} \sum_{\text{gram}_n \in s} \text{Count}_{\text{match}}(\text{gram}_n)}{\sum_{s \in \text{Ref Summaries}} \sum_{\text{gram}_n \in s} \text{Count}(\text{gram}_n)}
\end{equation}

\subsubsection{Perplexity}

\begin{equation}
\text{Perplexity} = \exp \left( -\frac{1}{N} \sum_{i=1}^{N} \log P(w_i \mid w_{i-1}) \right)
\end{equation}

\textbf{Features}: Used to evaluate the performance of language models. Indicates how well the model predicts the given text. Here, P is the conditional probability, and N is the length of the text.
Perplexity is an indicator used to evaluate the performance of language models. It measures how well the model predicts the given text.

\subsubsection{METEOR }

BLEU score has problems of The Lack of Recall, Use of Higher Order N-gram, Lack of Explicit Word-matching Between Translation and Reference, Use of Geometric Averaging of N-gram. The METEOR score is a metric designed to solve the problem of the BLEU score mentioned above\cite{banerjeeLAVIE2005meteor}. The METEOR score calculates a score based on explicit words and word matches between machine translation and reference translation. If multiple references are given, each score is scored and the highest score is used. The overall metric score for the system is calculated based on aggregate statistics accumulated with the entire test set. In summary, it is similar to BLEU, but it is a metric with added steps to consider synonyms and compare word stems.

\subsubsection{Test Case}
To evaluate the model, test cases were created. The test case structure is as follows:
\begin{figure}[h]
    \centering
    \includegraphics[width=1\linewidth]{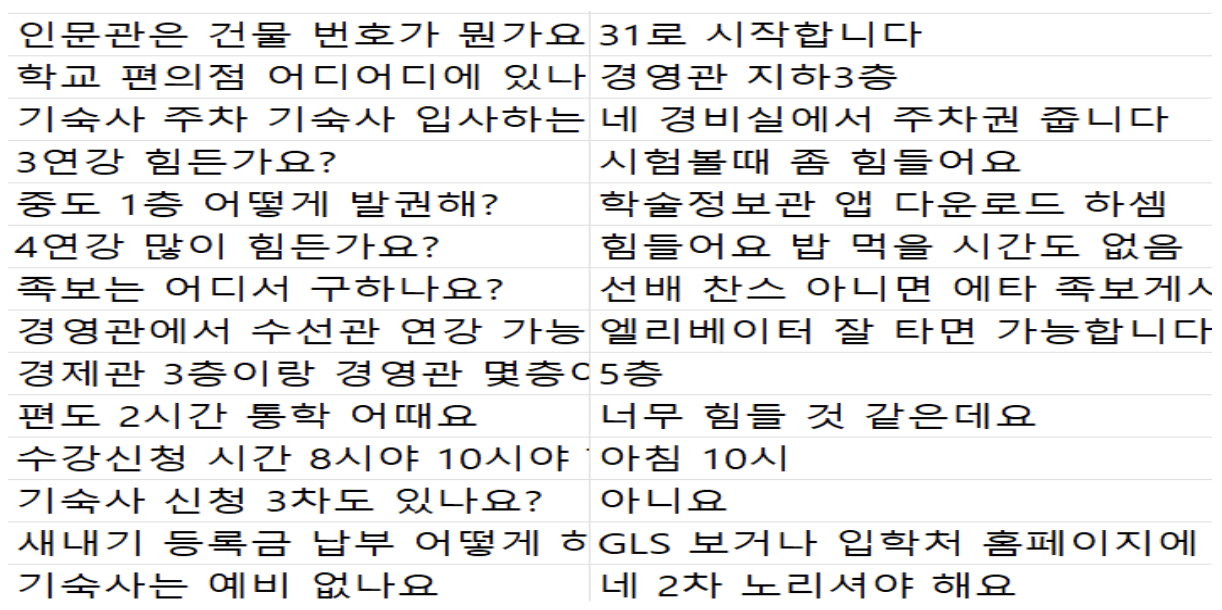}
    \caption{Test Case}

\end{figure}

\begin{table*}[!t]
\centering
\begin{tabular}{|m{2cm}|m{2cm}|m{6cm}|m{5cm}|}
\hline

Category                   & Tools             & Positives                                                                 & Negatives                                                                        \\
\hline
Server                     & Flask             & Flask does not require an expensive server. Can be modified on local computer & Time taken to receive responses is longer than a dedicated server.               \\
\hline
Chatbot                    & GPT-Based (Gpt-4o) & Plenty of data needed for training is already pre-trained.                & Needs tuning for training (costly).                                              \\
\hline
\multirow{2}{*}{Embedding} & KLUE/BERT-base    & Specialized for the Korean language dataset.                              & Relatively older technology compared to the latest transformer models (2021 release). \\
\cline{2-4}
                           & Ada v2            & Simplified API integration.                                               & Higher computational cost than KLUE/BERT-base.                                   \\
\hline
\end{tabular}
\caption{Previous research about University Chatbot system}
\end{table*}

\subsection{Result}
The performance of the LLM was evaluated using the BLEU, Perplexity, ROUGE, and METEOR metrics. The results of these evaluations are summarized as follows:

\textbf{BLEU Score}.
The BLEU (Bilingual Evaluation Understudy) score obtained was 0.0551. A BLEU score closer to 1 indicates a higher quality of machine translation, where the generated text closely matches the reference text. In this case, the relatively low BLEU score suggests that there is considerable room for improvement in generating responses that more closely align with the reference answers.


\textbf{Perplexity}. 
The Perplexity score obtained was 15.8756. Perplexity measures how well the language model predicts the next word in a sequence. A lower Perplexity score indicates that the model is better at predicting the text. In this context, a Perplexity score of 15.8756 suggests that the model is reasonably good at predicting the text but still has potential for enhancement to achieve a lower Perplexity, indicating better predictive performance.

\textbf{ROUGE Score}. The ROUGE (Recall-Oriented Understudy for Gisting Evaluation) scores obtained were as follows:
ROUGE-1: Recall: 0.0360, Precision: 0.0121, F-score: 0.0155
ROUGE-2: Recall: 0.0134, Precision: 0.0028, F-score: 0.0036
ROUGE-L: Recall: 0.0340, Precision: 0.0118, F-score: 0.0151
ROUGE scores evaluate the overlap of n-grams between the generated and reference texts. ROUGE-1 focuses on unigram overlap, ROUGE-2 on bigram overlap, and ROUGE-L on the longest common subsequence. 

\textbf{METEOR Score}.
The METEOR score obtained was 0.0243.

\section{Conclusion}

In this study, we developed the first chatbot for university freshmen, which utilizes community data to provide information that is more relevant to real life than the official materials provided by the school. This is expected to greatly help the social adaptation of new generations, especially those who prefer message-based communication, such as the Generation Z.

During the data collection process of this study, we identified major challenges, particularly the inability to distinguish between comments and replies, which prevented us from including additional questions in the question portion of the dataset. We decided to remove questions with question marks from comments, but this may have resulted in a loss of useful information. Future research should discuss how to handle the inclusion of additional questions in comments.

In our performance evaluation, we utilized metrics such as BLEU and ROUGE that have been used in previous studies, but we found that these metrics are not optimized for chatbot evaluation. Testing with human evaluators would have been ideal, but due to copyright issues and resource limitations, we were unable to do so. In addition, we believe that the difference between the length of the answers in the test cases and the length of the answers generated by the model may have contributed to the poor performance. In the future, we may consider utilizing test cases with answers of different lengths or evaluating them through surveys. We plan to ask the 'Everytime' platform to deploy the chatbot and get feedback from real users.

To clarify the limitations and strengths of the study, we would like to organize the limitations and strengths in a table format, which will serve as an important basis to guide the direction of future research. In addition, this study will be useful not only for university students but also for the broader Generation Z, which represents a very important step forward in modern society with a high preference for message-based communication.

{\small
\bibliographystyle{ieee_fullname}
\bibliography{egbib}
}

\end{document}